\documentclass{article}
\usepackage{ijcai22}

\usepackage{times}
\usepackage{soul}
\usepackage{url}
\usepackage[hidelinks]{hyperref}
\usepackage[utf8]{inputenc}
\usepackage[small]{caption}
\usepackage{graphicx}
\usepackage{amsmath}
\usepackage{amsthm}
\usepackage{booktabs}
\usepackage{algorithm}
\usepackage{algorithmic}
\usepackage{xcolor}
\usepackage{subcaption}
\usepackage{enumitem}
\title{Policy-Value Alignment and Robustness in Search-based Multi-Agent Learning}

\author{
Niko A. Grupen$^1$
\and
Michael Hanlon$^1$\and
Alexis Hao$^1$\and
Daniel D. Lee$^1$\and
Bart Selman$^1$
\affiliations
$^1$Cornell University\\
\emails
}
\begin{document}

\maketitle

\begin{abstract}
    Large-scale AI systems that combine search and learning have reached super-human levels of performance in game-playing, but have also been shown to fail in surprising ways. The brittleness of such models limits their efficacy and trustworthiness in real-world deployments. In this work, we systematically study one such algorithm, AlphaZero, and identify two phenomena related to the nature of exploration. First, we find evidence of policy-value misalignment---for many states, AlphaZero's policy and value predictions contradict each other, revealing a tension between accurate move-selection and value estimation in AlphaZero's objective. Further, we find inconsistency within AlphaZero's value function, which causes it to generalize poorly, despite its policy playing an optimal strategy. From these insights we derive VISA-VIS: a novel method that improves policy-value alignment and value robustness in AlphaZero. Experimentally, we show that our method reduces policy-value misalignment by up to 76\%, reduces value generalization error by up to 50\%, and reduces average value error by up to 55\%.
\end{abstract}

\section{Introduction}
Learning algorithms that combine self-play reinforcement learning and search have grown in popularity in recent years due to their ability to reach superhuman levels of performance in combinatorially complex environments~\cite{silver2017alphago}. More recently, an examination of AlphaZero \cite{silver2017alphazero} revealed that it obeys neural scaling laws with respect to play strength and network capacity~\cite{neumann2022scaling}---further suggesting that hybrid algorithms, as in other domains (e.g. large language models~\cite{kaplan2020scaling}), may have significant room to scale further. As these algorithms continue to scale towards real-world complexity, it calls into question whether state-of-the-art hybrid systems are ready for real-world deployment.

To date, researchers and practitioners have a very limited understanding of why the neural networks underpinning these hybrid algorithms make the decisions that they do---consider the infamous ``move 37" in AlphaGo vs. Lee Sedol~\cite{move37alphago}---and what they might learn about the environment and other agents within it. Though significant progress has been made in the area of concept-based interpretability, where post-hoc analysis has shown that many human-understandable concepts can be accurately regressed from AlphaZero's representations \cite{mcgrath2022acquisition}, opaque artifacts of self-play learning still remain. For example, studies leveraging adversarial game-play have shown that AlphaZero is surprisingly susceptible to minor perturbations in opponent behavior \cite{lan2022alphazero} and can be exploited to lose games that could be easily won by amateur human players \cite{wang2022adversarial}.
\begin{figure*}[t!]
    \centering
    \includegraphics[width=0.9\textwidth]{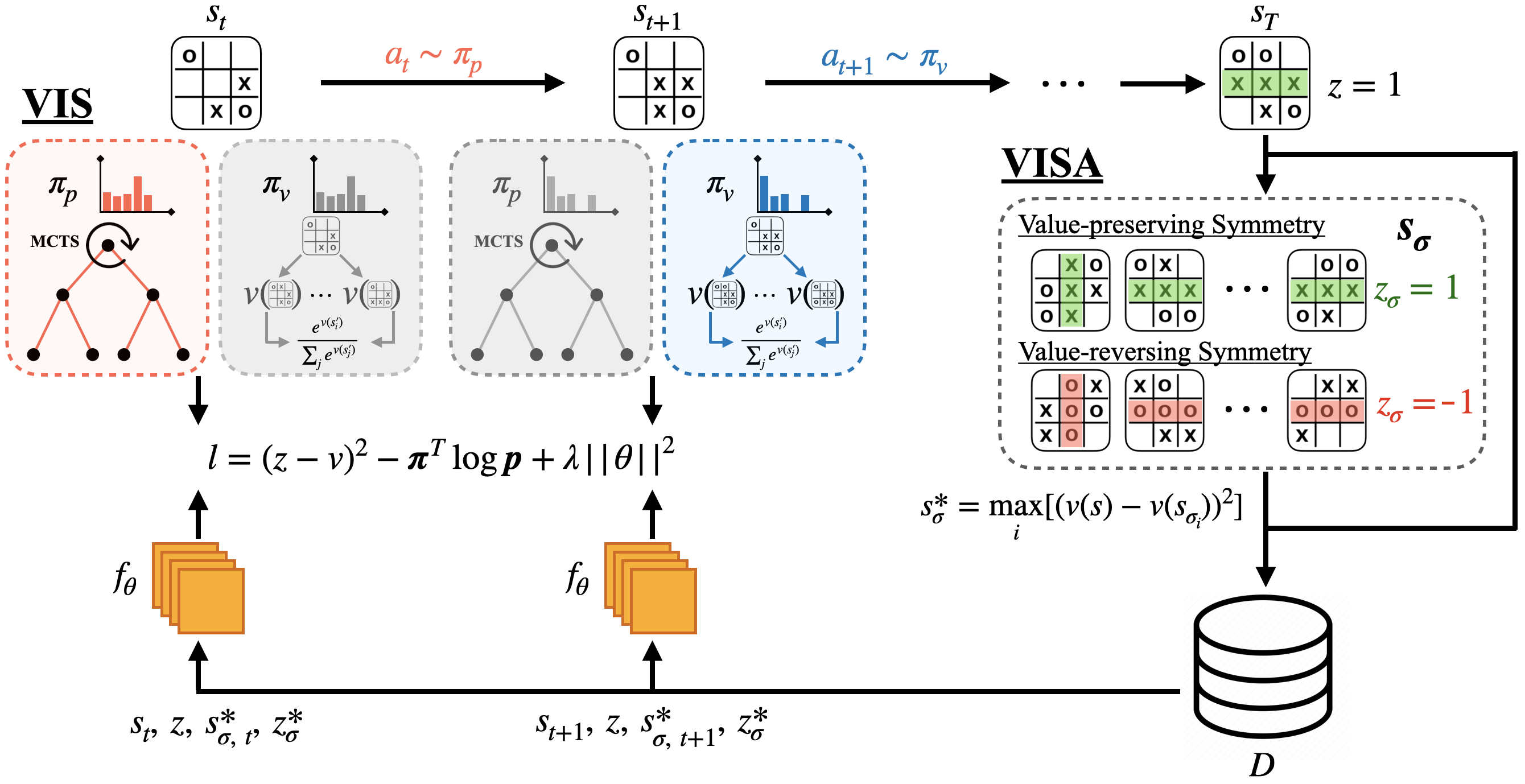}
    \caption{An overview of our proposed method. VISA-VIS combines two improvements to the AlphaZero algorithm: (i) \textbf{Value-Informed Selection (VIS)}, where action selection alternates stochastically between sampling actions from MCTS search probabilities ($a_t \sim \pi_p$) and sampling from a set of action probabilities derived from AlphaZero's value function ($a_t \sim \pi_v$); and (ii) \textbf{Value-Informed Symmetric Augmentation (VISA)}, a form of data augmention that targets uncertainty in AlphaZero’s value function. Specifically, before adding a state $s_t$ to the replay buffer $D$, VISA generates a set of symmetric state transformations $\boldsymbol{s_\sigma}$, evaluates each transformed state $s_{\sigma_i}$ with its value function, and returns the state $s^*_\sigma$ that differs most from AlphaZero's value estimate for the original state $s_t$. Both $s_t$ and $s^*_\sigma$ are added to $D$.}
    \label{fig:overview}
\end{figure*}

In this work, we discover two phenomena that are related to the nature of AlphaZero's exploration and may underlie its documented failure modes. First, we identify a novel emergent phenomenon, \textbf{policy-value misalignment}, in which AlphaZero's policy and value predictions contradict each other. Policy-value misalignment is a direct consequence of AlphaZero's loss function, which includes separate terms for policy and value errors individually, but no constraint on their consistency. We introduce an information-theoretic definition of policy-value misalignment and show how it can be measured for any AlphaZero network. Next, we find evidence of \textbf{inconsistency within AlphaZero's value function} with respect to symmetries in the environment in which it is trained. As a result, AlphaZero's value predictions generalize surprisingly poorly to infrequently-visited and unseen states. 

Crucially, we show these \textbf{phenomena are masked by search}, as it is possible for both to be present even when AlphaZero plays optimally within its search process. This work therefore demonstrates an interesting tension: search allows AlphaZero to achieve unprecedented scale and performance, but may also hide deficiencies in its learned components (i.e. its policy and value network).
% Moreover, AlphaZero's heavy reliance on search places an onus on its network to generalize well outside of the search distribution and is a possible avenue for the adverse effects of adversarial examples reported in prior work. 

To address these issues, we propose two modifications to AlphaZero aimed directly at policy-value alignment and value function consistency. First, we introduce Value-Informed Selection (VIS): a modified action-selection rule in which action probabilities are computed using a one-step lookahead with AlphaZero's value network. VIS then alternates stochastically between this value action selection rule and AlphaZero's standard policy (which is derived from search probabilities). During training, this forces alignment between AlphaZero's policy and value predictions. Next, we introduce Value-Informed Symmetric Augmentation (VISA): a data augmentation technique that takes into account uncertainty in AlphaZero's value network. Unlike data augmentation techniques that naively add states to a network's replay buffer, VISA considers symmetric transformations of a given state and stores only the transformed state whose value differs most from the network's value estimate for the original, un-transformed state.

VISA and VIS can improve both policy-value alignment and value robustness in AlphaZero simultaneously. We refer to this joint method as VISA-VIS. In experiments across a variety of symmetric, two-player zero-sum games, we show that VISA-VIS reduces policy-value misalignment by \textbf{up to 76\%}, reduces value generalization error by \textbf{up to 50\%}, and reduces average value prediction error \textbf{by up to 55\%}; as compared to AlphaZero.
\vspace{7pt}

\noindent \textbf{Contributions:}
In sum, we offer five primary contributions:
\begin{itemize}[leftmargin=*]
    \item We systematically study AlphaZero and identify two phenomena--policy-value misalignment and value function inconsistency--that are masked by AlphaZero's search.
    \item We define an information-theoretic measure of policy-value misalignment that can be used to evaluate any trained AlphaZero network.
    \item We propose Value-Informed Selection (VIS): A modified action-selection rule for AlphaZero that forces alignment between policy and value predictions.
    \item We propose Value-Informed Symmetric Augmentation (VISA): A data augmentation technique that leverages symmetry and value function uncertainty to learn a more consistent value network.
    \item We introduce a new method, VISA-VIS, that combines VIS and VISA into a single algorithm that extends AlphaZero. Experimentally, we show that VISA-VIS reduces policy-value misalignment by up to 76\%, reduces value generalization error by up to 50\%, and reduces average value predictions error by up to 55\% compared to vanilla AlphaZero.
\end{itemize}

\section{Related Work} 
The continued success of hybrid algorithms that combine search and self-play RL has drawn research interest into their robustness and interpretability.

\paragraph{Adversarial Examples}
In supervised learning settings, it is well-documented that both synthetic \cite{goodfellow2014explaining} and naturally-occuring \cite{hendrycks2021natural} adversarial examples can induce sub-optimal behavior from neural network-based classifiers. Recently, similar tactics have proven effective in identifying the weaknesses of RL agents. For example, it has been shown that small perturbations of an agent's observations can cause large decreases in reward \cite{huang2017adversarial,kos2017delving} and that performance degradation can be exacerbated by leveraging knowledge of an agent's behavior \cite{lin2017tactics}.

In multi-agent settings, RL agents are particularly vulnerable to adversarial behavior from other agents in the environment \cite{gleave2019adversarial}. Of particular relevance is the work of Wang et al.~\shortcite{wang2022adversarial}, which showed that a professional-level AlphaZero Go agent can be tricked into losing games by simple adversarial moves that are easily countered by amateur human players. Further studies have shown that AlphaZero is susceptible to other state- and action-space perturbation \cite{lan2022alphazero}. We posit that these surprising failure modes are a symptom of the phenomena we describe in this work---policy-value misalignment and value function inconsistency.

\paragraph{Interpretability}
RL interpretability methods are divided into those that improve the transparency of agent decision-making by either (i) introducing intrinsically-interpretable policy and value network architectures; or (ii) generating post-hoc explanations for previously-trained networks. Beyond more traditional approaches that are derived directly from supervised learning counterparts--such as saliency maps \cite{greydanus2018visualizing,selvaraju2017grad}---advances in reward-agnostic behavioral analysis \cite{omidshafiei2022beyond} and intrinsically-interpretable policy learning schemes \cite{grupen2022concept} have greatly improved the interpretability of an RL agent's decision-making (and are applicable to multi-agent settings). 

For large-scale AI systems such as AlphaZero, seminal work on post-hoc, concept-based interpretability has shown that many human-understandable concepts are encoded in AlphaZero's policy network during training \cite{mcgrath2022acquisition}. Additional studies have employed model probing and behavioral analysis to understand the concepts represented by AlphaZero for the game of Hex \cite{forde2022concepts}. These studies highlight the search-learning conundrum proposed in our work---it is possible for AlphaZero to learn complex playing strategies and sophisticated human-understandable concepts while simultaneously demonstrating inconsistencies in its policy and value predictions.

\section{Background}
\subsection{AlphaZero}
\label{sec:alphazero}
AlphaZero is a hybrid algorithm that combines self-play RL and Monte-Carlo Tree Search (MCTS). We review AlphaZero's search and learning processes below, as both are crucial to understanding our method.

\paragraph{Setup}
AlphaZero maintains a search tree in which each node is a state $s$ and each edge $(s, a)$ represents taking an action $a$ from $s$. Each edge is described by a tuple of statistics $\{N(s,a), W(s,a), Q(s,a), P(s,a)\}$, holding the edge's visit count, total action value, mean action value, and selection probability, respectively. AlphaZero uses a two-headed neural network $f_\theta$ with parameters $\theta$ to compute policy probabilities $\boldsymbol{p}$ and a value estimate $v$ for a state $s$ as follows:
\begin{equation}
    \label{eqn:az_network}
    (\boldsymbol{p}, v) = f_\theta(s)
\end{equation}
MCTS uses $f_\theta$ to evaluate states during the search process. 

\paragraph{Search}
In each state $s$, AlphaZero executes $n$ simulations of MCTS tree search. In each simulation, MCTS traverses the game tree by selecting edges that maximizes the upper confidence bound $Q(s,a)+U(s,a)$, where:
\begin{equation}
    \label{eqn:az_puct}
    U(s,a) = cP(s,a)\frac{\sqrt{\sum_b N(s,b)}}{1 + N(s,a)}
\end{equation}
encourages the exploration of lesser-visited states (within the search tree); and $c$ is a coefficient determining the level of exploration. When a leaf node $s_L$ is encountered, it is expanded, evaluated by AlphaZero's network to produce $(\boldsymbol{p}_L, v_L) \sim f_\theta(s_L)$, and initialized as follows:
\begin{equation}
    \label{eqn:az_node_init}
    N(s,a) = 0, W(s,a) = 0, Q(s,a) = 0, P(s,a) = p_{L, a}
\end{equation}
\noindent Then, in a backward pass up the search tree, the statistics of each traversed edge are updated as follows: $N(s,a) = N(s,a) + 1$, $W(s,a) = W(s,a) + v_L$, $Q(s,a) = W(s,a) / N(s,a)$.

\paragraph{Action Selection}
After MCTS, AlphaZero constructs a policy $\boldsymbol{\pi}$, where the probability $\pi(a|s)$ of selecting action $a$ from state $s$ is computed directly from visitation counts:
\begin{equation}
    \label{eqn:mcts_policy}
    \pi(a|s) = \frac{N(s,a)^{1/\tau}}{\sum_b N(s,b)^{1/\tau}}
\end{equation}
and $\tau$ is a temperature parameter that controls exploration. During both training and test-time rollouts, AlphaZero samples actions $a \sim \boldsymbol{\pi}$ according to this policy.

\paragraph{Learning}
Starting from an initial state $s_0$ (e.g. an empty board), AlphaZero proceeds with self-play, using the aforementioned search process to select actions until a terminal state $s_T$ is reached. A score $z$ is produced for $s_T$ according to the rules of the game and tuples $(s_t, a_t, z)$ are stored in a replay buffer $D$ for each time-step $t \in [0, T]$ in the trajectory. AlphaZero's neural network is updated to minimize the loss function:
\begin{equation}
    \label{eqn:az_loss}
    l = (z - v)^2 - \boldsymbol{\pi}^T \log \boldsymbol{p} + \lambda||\theta||^2
\end{equation}
where $(\boldsymbol{p}, v) \sim f_\theta(s)$ as in Equation \eqref{eqn:az_network}, $\boldsymbol{\pi}$ follows from Equation \eqref{eqn:mcts_policy}, and $\lambda$ is a coefficient weighting $L2$-regularization. Intuitively, Equation \eqref{eqn:az_loss} encourages the network to (i) minimize the error between predicted values $v$ and game outcomes $z$; and (ii) maximize similarity between predicted policy probabilities $\boldsymbol{p}$ and search probabilities $\boldsymbol{\pi}$.

\section{Method}
\label{sec:visa_vis}
In this section, we introduce the proposed VISA-VIS method that combines two modifications to the AlphaZero algorithm: Value-Informed Selection (VIS) and Value-Informed Symmetric Augmentation (VISA). We describe each in detail in the subsections below, and an overview of our proposed approach is presented in Figure \ref{fig:overview}.

\subsection{Value-Informed Selection}
\label{sec:vis}
Recall AlphaZero's loss function from Equation \eqref{eqn:az_loss}. Though this objective incentivizes accurate policy and value predictions individually, it imposes no constraint on their consistency. It is therefore possible for AlphaZero's network to output policy probabilities $\boldsymbol{p}$ that contradict its value prediction $v$---e.g. during inference, $\boldsymbol{p}$ places large probability mass on a winning action while $v$ predicts that the current state is a losing one. We find that this occurs frequently in practice. To enforce policy-value consistency, we introduce Value-Informed Selection (VIS). VIS is a novel action selection rule in which AlphaZero alternates between selecting actions with its policy (as informed by the search probabilities in Equation \eqref{eqn:mcts_policy}) and selecting actions with its value function.

More formally, let $\boldsymbol{s'}$ be the set of successor states from a given state $s$, where each successor $s'_a \in \boldsymbol{s'}$ is the result of taking a legal action $a$ from $s$. We then define $\pi_v$ to be the action probabilities obtained by computing a one-step value lookahead from $s$:
\begin{equation}
    \pi_v(a|s) = \frac{e^{v(s'_a)}}{\sum\limits_{s'_b \in \boldsymbol{S'}} e^{v(s'_b)}}
\end{equation}
where $v(s)$ is the value of state $s$, as predicted by AlphaZero's value network. Renaming the search policy defined in Equation \eqref{eqn:mcts_policy} to be $\pi_p$, VIS then defines the action selection rule:
\begin{equation}
  \pi_{\textrm{VIS}}(a|s) =
    \begin{cases}
      \pi_p(a|s) & \text{if} \; \; \eta < \epsilon\\
      \pi_v(a|s) & \text{otherwise.}\\
    \end{cases}
\end{equation}
where $\eta {\sim} U(0,1)$ is drawn uniformly and $\epsilon$ is a threshold determining the likelihood of policy vs. value action selection. When used in conjunction with Equation \eqref{eqn:az_loss}, VIS forces AlphaZero's policy and value predictions to be aligned by swapping $\boldsymbol{\pi_p}$ and $\boldsymbol{\pi_v}$ stochastically. Intuitively, VIS pushes AlphaZero to align its policy and value function because it never knows which it will have to use for any given action selection.

\subsection{Value-Informed Symmetric Augmentation}
\label{sec:visa}
Exhaustively searching a combinatorial state-space is impractical, so during training, AlphaZero relies upon MCTS to reduce the set of states it encounters. However, this also places the onus on AlphaZero's network to generalize to states that are skipped by MCTS during training. We find that AlphaZero's value predictions generalize poorly to such states.

To improve the generalization capabilities of AlphaZero's value network, we propose Value-Informed Symmetric Augmentation (VISA): a form of data augmention that explicitly targets uncertainty in AlphaZero's value function. For each state $s$ encountered during training, VISA first generates a set $\boldsymbol{s_\sigma} = \{s_{\sigma_1}, ..., s_{\sigma_N}\}$ of $N$ additional states, where each $s_{\sigma_i} \in \boldsymbol{s_\sigma}$ is the result of applying a symmetric transformation $\sigma_i(s)$ to $s$. VISA then finds the transformed state $s^*_\sigma$ with maximum value uncertainty as:
\begin{equation*}
    s^*_\sigma = \max_i[(v(s) - v(s_{\sigma_i}))^2]
\end{equation*}
and adds both $s$ and $s^*_\sigma$ to AlphaZero's replay buffer. Note that, by adding only $s^*_\sigma$, VISA is an improvement over both random data augmentation---it targets the augmentation for which the value network is least certain---and exhaustive data augmentation---it is more memory efficient than storing all of $\boldsymbol{s_\sigma}$ in the replay buffer.

In this work, we consider three principal symmetric transformations: rotation and reflection from the dihedral group, and inversion. Intuitively, in the board game setting (as in Figure \ref{fig:overview}), these transformations correspond to rotating the game board, mirroring the game board (vertically or horizontally), or flipping each player's pieces. Importantly, we note that board inversion also requires inverting the ground-truth game-tree value $z$ when computing Equation \eqref{eqn:az_loss}.

\subsection{Measuring Policy-Value Misalignment}
We also introduce an information-theoretic measure of policy-value misalignment. Given AlphaZero's standard policy $\pi_p$ and the value-informed policy $\pi_v$ from Section \ref{sec:vis}, it is possible to compute policy-value misalignment as the KL-divergence:
\begin{equation}
    \label{eqn:misalignment}
    D_{\textrm{KL}}(\pi_p || \pi_v)
\end{equation}
\noindent Intuitively, a model with strong policy-value alignment will yield a lower divergence term, whereas policy and value networks that are inconsistent will result in higher divergence.

\section{Experiments}
\begin{figure*}[t!]
    \centering
    \includegraphics[width=0.95\textwidth]{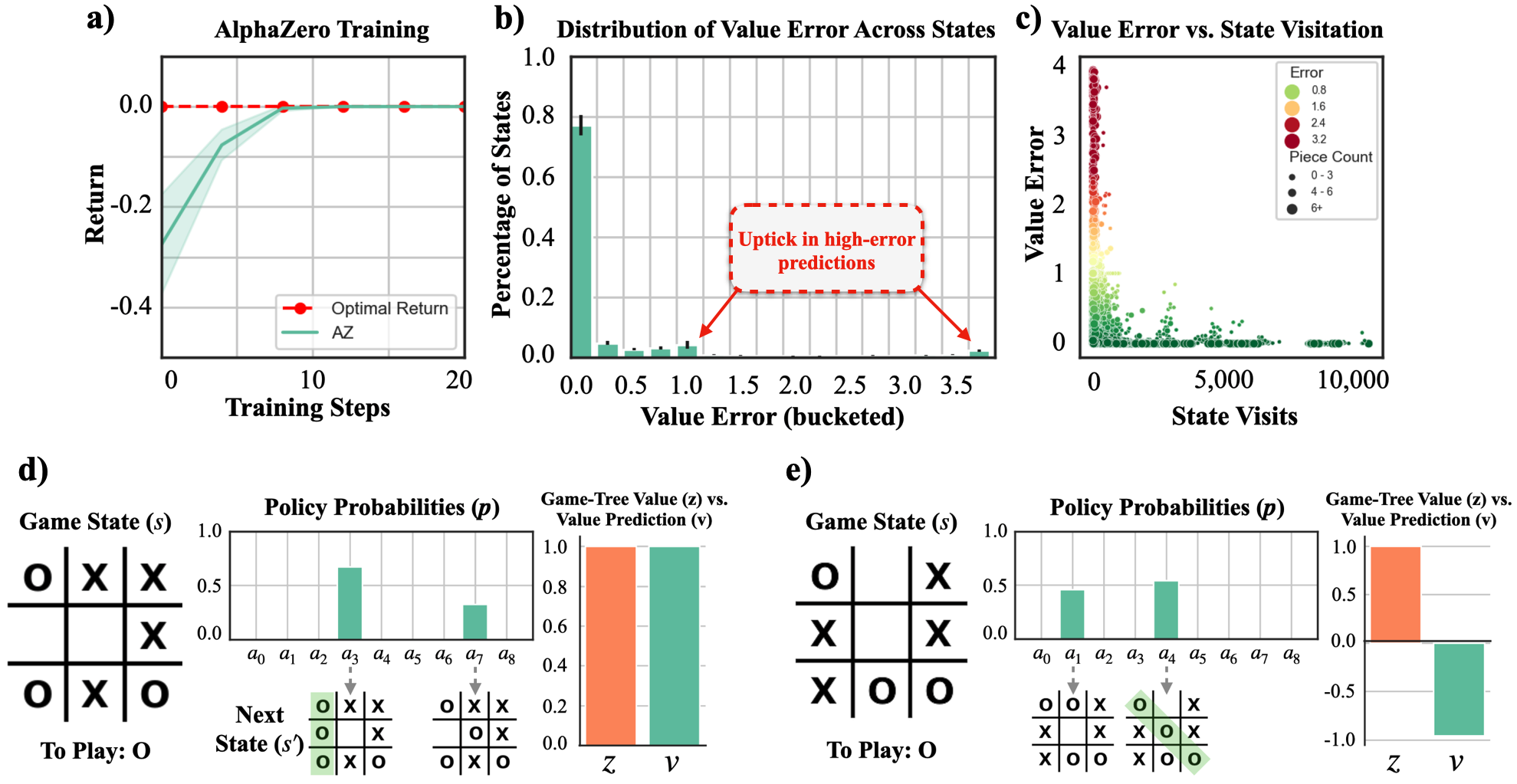}
    \caption{Evaluation of AlphaZero in an enumerable game tree. \textbf{a)} AlphaZero performance during training. For a simple game, AlphaZero quickly converges to an optimal strategy. \textbf{b)} Despite playing optimally, an exhaustive analysis of AlphaZero's value function performance shows that it makes many incorrect and high-error predictions (relative the true game-tree value for each state). \textbf{c)} Computing value error as a function of state visitation reveals that AlphaZero's value function is failing to generalize to infrequently visited and unseen states. \textbf{d)} A qualitative example where AlphaZero's policy and value predictions are both correct. \textbf{e)} A qualitative example demonstrating policy-value misalignment---AlphaZero's policy prediction picks a winning move, but its value prediction is incorrect (predicts a loss).}
    \label{fig:3x3_results}
\end{figure*}
In this section, we first demonstrate the presence of policy-value misalignment and value function inconsistency in AlphaZero. Then, we evaluate VISA-VIS and examine the extent to which it alleviates these phenomena.

\paragraph{Environments:}
We consider three solved, symmetric two-player zero-sum games of increasing complexity: Tic-Tac-Toe, 4x4 Tic-Tac-Toe, and Connect Four. We emphasize the importance of studying solved games to our analysis. In order to properly measure the scope of AlphaZero's value generalization error, we need a suitable ground truth with which to compare AlphaZero's predictions. Solved games fulfill that need. In each game, the outcome can take on a value of $z=1, 0,$ and $-1$ for wins, losses, and draws, respectively.

\paragraph{Training Details:}
AlphaZero is trained for a fixed number of time-steps following the algorithm in Section \ref{sec:alphazero}. AlphaZero's network consists of a ResNet \cite{he2016deep} backbone with separate heads for policy and value prediction. All details are the same for VISA-VIS training runs, but following the modifications outlined in Section \ref{sec:visa_vis}. Additional details, including hyperparameters, are outline in Appendix \ref{apdx:training}.

\subsection{AlphaZero in an Enumerable Game Tree}
We first present a systematic study of AlphaZero's performance in an enumerable game-tree: Tic-Tac-Toe. Due to the small state-space of Tic-Tac-Toe, it is possible to evaluate AlphaZero's predictions against every legal state, providing us with a holistic view of AlphaZero's behavior.

After training, the performance of each AlphaZero model is verified in two ways: First, we play each model against an oracle opponent for $1,000$ matches across five random seeds and score each match according to the rules of the game ($z \in \{1, 0, -1\}$). For completeness, we include both games in which AlphaZero moves first, and those in which AlphaZero moves second. Next, we verify the accuracy of AlphaZero's value predictions by iterating through the legal state-space and computing the mean-squared error between AlphaZero's value predictions and the true game-tree value. The results of this analysis are presented in Figure \ref{fig:3x3_results}. 

From the training curves in Figure \ref{fig:3x3_results}a, it is clear that AlphaZero converges to a strong game-playing strategy very quickly---the expected outcome is a tie under optimal play---which is expected. The result of the value prediction analysis, on the other hand, is quite surprising. In Figure \ref{fig:3x3_results}b, we find that, despite AlphaZero's ability to play optimally, its value predictions are not always accurate. In fact, when the error of AlphaZero's value predictions is quantized, there is a noticeable error bump when $e > 1.0$---indicating AlphaZero is predicting a loss when the true game-tree value is a win (or vice versa) and another error bump when $e > 3.5$---indicating that AlphaZero is predicting a loss \textit{very confidently} when the true game-tree value is a win (or vice versa).

When plotted as a function of the number of visits made by AlphaZero to each state during training (see Figure \ref{fig:3x3_results}c), it becomes clear that the error of AlphaZero's value predictions is highest for infrequently-visited or unseen states, indicating that AlphaZero has failed to generalize. Moreover, as shown in the qualitative examples from Figure \ref{fig:3x3_results}e, which includes both AlphaZero's policy predictions and its corresponding value estimates, we find that, in many instances, AlphaZero's policy predictions are correct (they choose the winning move), despite value predictions that are wildly inaccurate (they predict a loss, despite the winning opportunity).

Altogether, this analysis provides evidence that (i) AlphaZero's value predictions are inconsistent and generalize poorly; and (ii) AlphaZero's value predictions are not aligned with its policy predictions (i.e. AlphaZero suffers from policy-value misalignment).

\subsection{Ablation: VISA, VIS, and VISA-VIS}
Next, we investigate how well our proposed method, VISA-VIS, corrects AlphaZero's susceptibility to generalization error and policy-value misalignment. We perform the same test-time analysis over trained VISA-VIS models and, to gauge the contributions of each component of our method, perform an ablation over both VISA and VIS separately. Aside from AlphaZero and VISA-VIS, we report results for an AlphaZero model that is trained from random initial states (each training rollout begins from a random legal, non-terminal state). Though training with random initial states does not scale to more complex games, it provides a lower-bound for value error in this setting (i.e. how well can AlphaZero do when it is allowed to visit all states many times).

\begin{figure}[t!]
    \centering
    \makebox[\linewidth][c]{\includegraphics[width=0.9\linewidth]{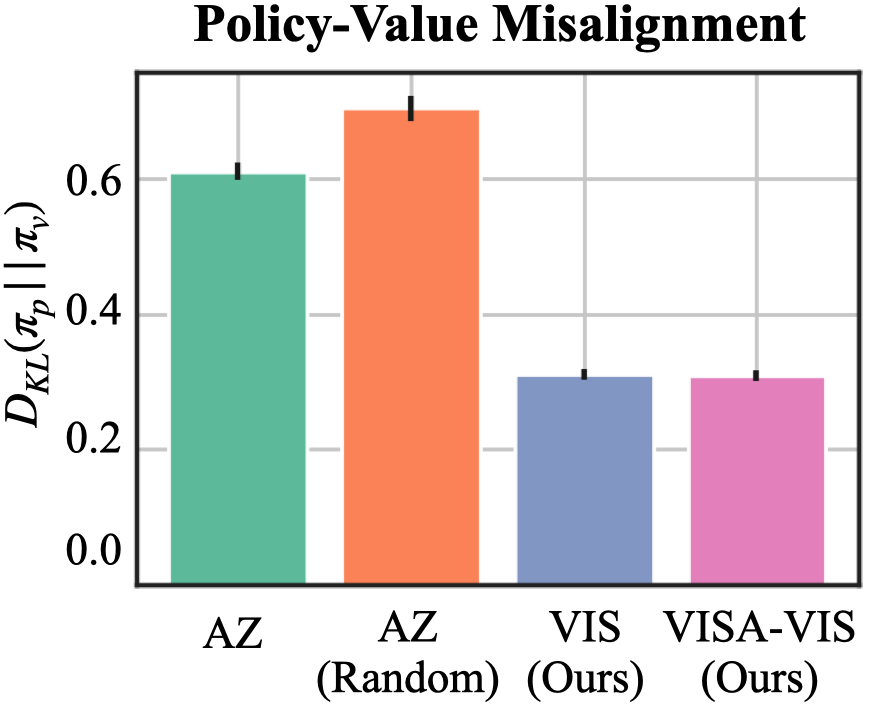}}
    \caption{Policy-value misalignment of each method, as given by the information-theoretic measure introduced in Equation \ref{eqn:misalignment} (lower is better). VIS reduces policy-value misalignment by over 50\% over AlphaZero (AZ) and maintains alignment when combined with VISA (VISA-VIS).}
    \label{fig:ablation_misalignment}
\end{figure}

\paragraph{Policy-Value Misalignment}
First, we compute the average policy-value misalignment exhibited by each method across all environmental states, as defined by Equation \eqref{eqn:misalignment}. As shown in Figure \ref{fig:ablation_misalignment}, VIS, and therefore VISA-VIS, significantly improve the consistency of the model's policy and value predictions, reducing policy-value misalignment by 50\% compared to AlphaZero and by over 60\% as compared to AlphaZero Random. These results also provide a great example of how value prediction accuracy and policy-value misalignment differ in nature. Consider AlphaZero Random, for example. Despite observing all game states directly, AlphaZero Random exhibits a high policy-value misalignment. This is because value estimation alone does not tell the whole story---it is still possible for AlphaZero's value function to be inconsistent with its policy.

\begin{figure}[t!]
    \centering
    \makebox[\linewidth][c]{\includegraphics[width=0.95\linewidth]{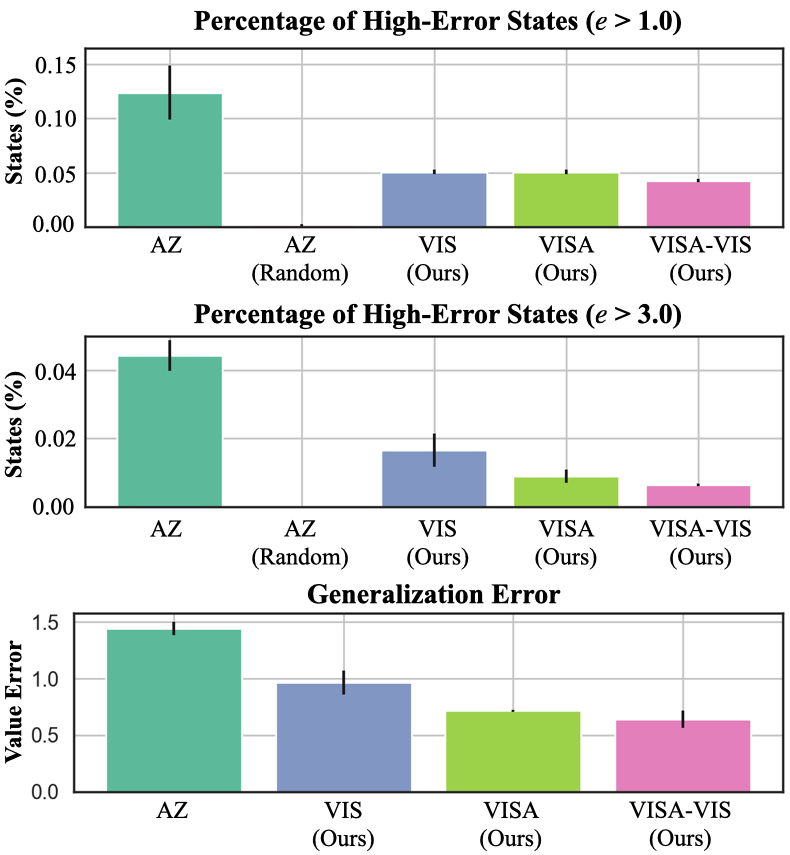}}
    \caption{Value error for each method (lower is better). VISA-VIS significantly reduces the percentage of medium ($e > 1.0$) and high error ($e > 3.0$) value predictions and reduces the average generalization error of AlphaZero's value network by over 50\%.}
    \label{fig:ablation_error}
\end{figure}

\paragraph{Value Error}
Value error results for this ablation are reported in Figure \ref{fig:ablation_error}. There we see that VISA, VIS, and VISA-VIS each significantly improve the quality of AlphaZero's value predictions. In particular, VISA-VIS reduces the percentage of states with a medium value error ($e > 1.0$) by almost 10\% as compared to AlphaZero and produces a high value error prediction ($e > 3.0$) for only 0.6\% of states ($6.9\times$ fewer than AlphaZero), which is extremely close to the perfect score obtained by the version of AlphaZero that observes all states (AZ Random). In terms of generalization, we find that VISA-VIS reduces average value generalization error by over $50\%$ (from an average of 1.44 to 0.64). Notably, for both high-error predictions and generalization error, we see that the performance improvements of VISA and VIS are complementary---VISA and VIS both decrease the number high-error states, but value error is decreased the most when they are combined (and likewise for generalization error).

\subsection{Scaling Up}
\begin{figure*}[t!]
    \centering
    \includegraphics[width=0.95\textwidth]{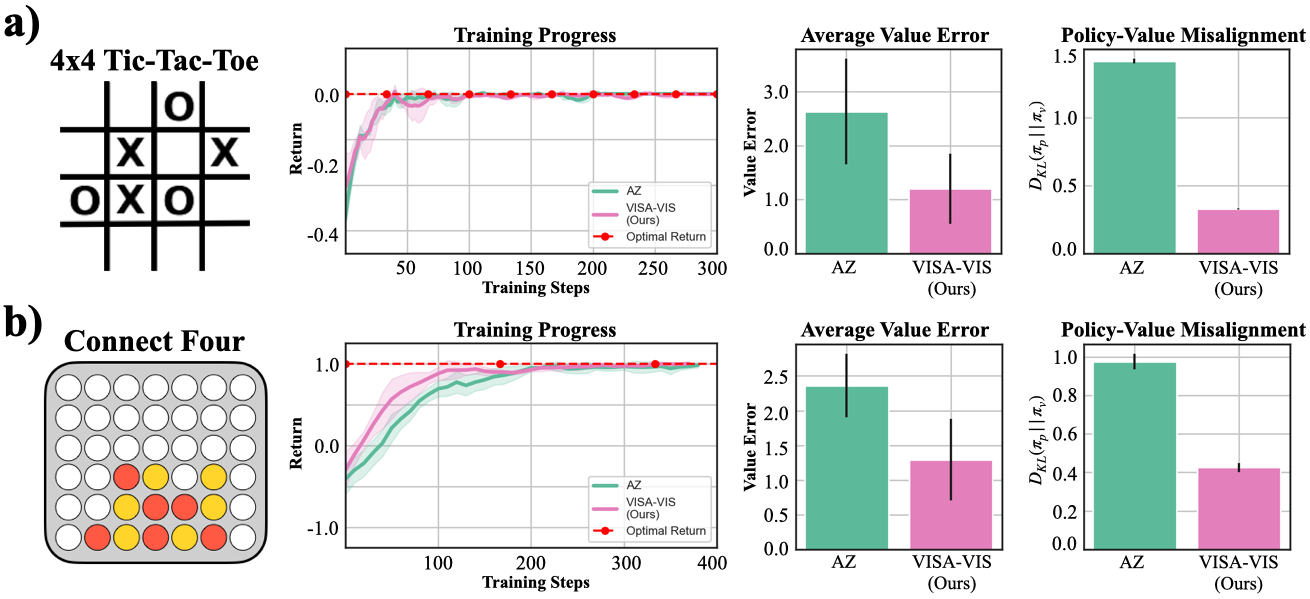}
    \caption{AlphaZero vs. VISA-VIS in more complex environments. \textbf{a)} In 4x4 Tic-Tac-Toe, An adversarial state analysis reveals over 10,000 states in which AlphaZero performs poorly (high value error, high policy-value misalignment). VISA-VIS largely corrects AlphaZero's errors in these states, reducing average value error by 55\% and policy-value misalignment by 76\%. Moreover, During training, AlphaZero and VISA-VIS converge at a similar rate, indicating that VISA-VIS attains with no corresponding increase in sample complexity. \textbf{b)} In Connect Four, the adversarial analysis yields qualitatively similar results. AlphaZero again demonstrates value function inconsistency---average value error of 2.36 across adversarial states---and policy-value misalignment---average KL-divergence of 0.975. These errors are again reduced by VISA-VIS, which averages a value error of 1.29 and a policy-value misalignment score of 0.425 over the same adversarially-generated states (representing a 45\% and 56.5\% reduction, respectively). Moreover, in terms of sample complexity, VISA-VIS converges more quickly than AlphaZero on average (in the Connect Four game). This reveals an additional benefit of VISA-VIS---symmetric augmentation reduces the time it takes to learn an optimal policy.}
    \label{fig:complex_results}
\end{figure*}

Here we demonstrate the scalability of our method through evaluation in significantly more complex games. The results of this analysis are presented in Figure \ref{fig:complex_results}.

\paragraph{Adversarial State Detection}
In more complex environments, it is not possible to enumerate the state-space and evaluate each state individually. We therefore make two modifications to our evaluation strategy to support more complex games. First, we focus primarily on endgame states. Endgame states allow us to target AlphaZero's most important decisions---a single move may determine the game's outcome---while still allowing us to extract a ground truth outcome from the game-tree, which we can use to verify AlphaZero's predictions.

Second, inspired by prior work on adversarial examples for AlphaZero \cite{wang2022adversarial}, we define an adversarial strategy that aids in uncovering states for which AlphaZero makes inconsistent predictions. Specifically, at test-time, after computing search-informed action probabilities with Equation \eqref{eqn:mcts_policy}, we force AlphaZero to take the action $a_t = \min_a[\pi(a|s)]$ with minimum assigned probability. Because $\pi$ encodes the expected value of taking action $a$ from state $s$, we are effectively forcing AlphaZero to play the worst move available. During each rollout, this adversarial strategy nudges AlphaZero into subsets of the state-space that are most likely to have been overlooked by search during training.
\begin{table}
  \caption{Adversarially-detected States.}
  \label{tab:adversarial_states}
  \centering
  \begin{tabular}{lll}
    \toprule
    Game     & Adversarial States     & Avg. Value Error\\
    \midrule
    4x4 Tic-Tac-Toe & \hspace{20pt}10,241 & \hspace{20pt}2.63 \\
    Connect Four & \hspace{20pt}52,768 & \hspace{20pt}2.36 \\
    \bottomrule
  \end{tabular}
\end{table}
This simple adversarial strategy proves effective at generating states that are poorly evaluated by AlphaZero. In each of our environments, we played a trained AlphaZero model against itself for $100,000$ matches using the aforementioned adversarial detection strategy and collected all unique endgame states for which AlphaZero made a high-error ($e > 1.0$) value prediction. In 4x4 Tic-Tac-Toe, the adversarial detector uncovered over $10,000$ states that received high-error value estimates \textbf{in each of three separately trained AlphaZero models}. The detector similarly uncovered over $50,000$ states for Connect Four. A summary of these statistics is available in Table \ref{tab:adversarial_states}.

In Figure \ref{fig:complex_results}a, which shows the results for our adversarial analysis in 4x4 Tic-Tac-Toe, we see that regular AlphaZero displays both poor value function consistency---with an average value error of 2.63 for these states---and significant policy-value misalignment---with average KL-divergence $D_{KL}(\pi_p || \pi_v) = 1.41$. Our method VISA-VIS corrects both of these issues, decreasing average value error 55\% to 1.20 and reducing policy-value misalignment by 76\% down to $D_{KL}(\pi_p || \pi_v) = 0.327$.

Figure \ref{fig:complex_results}b presents our adversarial analysis in Connect Four. Here we find a qualitatively similar result to the 4x4 Tic-Tac-Toe case. Our adversarial detector is able to identify over 50,000 states that yield high-error predictions from AlphaZero. As before, VISA-VIS corrects the high-error predictions. More specifically, the adversarial states result in an average value error of 2.36 and an average policy-value misalignment score of 0.975 from AlphaZero. VISA-VIS compares favorably with an average error of 1.29 and an average policy-value misalignment score of 0.425---representing a 45\% and 56.5\% reduction over AlphaZero, respectively.

Altogether, these results not only show that policy-value misalignment and value function inconsistency are general phenomena that span multiple environments, but they also provide evidence that our method, VISA-VIS, can reliably combat both phenomena simultaneously.

\paragraph{Sample Complexity}
We also plot the game-playing performance of AlphaZero and VISA-VIS throughout training. As in the 3x3 case, the expected outcome of 4x4 Tic-Tac-Toe is a tie under optimal play ($z=1$). As shown in Figure \ref{fig:complex_results}a, we find evidence that the time-to-convergence of AlphaZero and VISA-VIS are roughly the same. Overall, VISA-VIS is able to attain the aforementioned improvements in policy-value alignment and value functon consistency without a loss in sample complexity.

In the case of Connect Four, we see that VISA-VISA has a more significant positive impact on sample complexity. In Figure \ref{fig:complex_results}b), it is clear that VISA-VIS converges to an optimal strategy more quickly than AlphaZero on average (the expected outcome for the first player is a win in Connect Four). This result suggests that, in this more complex game, the symmetric data augmentation of VISA-VIS reduces the sample complexity needed to learn an optimal policy. This highlights yet another benefit of VISA-VISA.

\section{Conclusion}
In this work, we presented a systematic study of AlphaZero, which uncovered two phenomena---policy-value misalignment and value function inconsistency---that hurt AlphaZero's robustness. Crucially, both of these phenomena are masked by search, indicating that it is still possible for AlphaZero to play optimally despite their existence and may contribute to AlphaZero's susceptibility to adversarial examples. We introduced VISA-VIS, which combines Value-Informed Selection (VIS)---a novel action selection criteria for AlphaZero---and Value-Informed Symmetric Augmentation (VISA)---a novel data augmentation that targets value function uncertainty with respect to symmetry. Together, VISA-VIS leads to significant improvements for both policy-value alignment and value function robustness in AlphaZero, as demonstrated in our experiments.

\section*{Ethical Statement}
This work examines the robustness and reliability of AlphaZero. As large-scale AI systems like AI proliferate in real-world use cases, it is important to understand their limitations and potential failure modes. Algorithms that leverage self-play reinforcement learning are particularly susceptible to potential harms (e.g. safety, reliability, interpretability), because they do not observe any human-provided data during training. We posit that, in order to ensure that such systems are generally helpful, and to mitigate unforeseen or unintended harms during deployment, we must ensure their robustness to potential attacks (e.g. adversarial inputs) and overall reliability. The goal of this work was to take a small step in that direction. This work does not make use of sensitive data or include as part of its results a sensitive task or study.

% \section*{Acknowledgments}

%% The file named.bst is a bibliography style file for BibTeX 0.99c
\bibliographystyle{named}
\bibliography{bib}

\clearpage
\appendix

\section{Training Details (cont'd)}
\label{apdx:training}
Here we provide additional training details, including the search parameters, training hyper parameters, and other learning details that are used when training AlphaZero and VISA-VIS.

\paragraph{Training parameters}
Both algorithm's were trained for $500,000$ games, $1.75$M games, $1.75$M games, and $7.5$M games for Tic-Tac-Toe, 4x4 Tic-Tac-Toe, and Connect Four, respectively. The joint policy-value network was represented by a ResNet backbone with separate heads outputting policy and value predictions, as in Equation \eqref{eqn:az_network}. Each fully-connected layer was initialized with a width of $128$ neurons and the depth of the network backbone was scaled to account for game complexity----$d=2$ ResNet modules for Tic-Tac-Toe, $d=4$ for 4x4 Tic-Tac-Toe and Connect Four. Batch size and learning rate also varied for each game: (i) Tic-Tac-Toe used a learning rate of $1\mathrm{e}{-}3$ and batch size of $64$; (ii) 4x4 Tic-Tac-Toe used a learning rate of $1\mathrm{e}{-}4$ and batch size of $128$; and (iii) Connect Four used a learning rate of $1\mathrm{e}{-}4$ and batch size of $256$. A coefficient of $\lambda=1\mathrm{e}{-}4$ was used to weight L2-regularization in all games, as described in Equation \eqref{eqn:az_loss}.

\begin{table}
  \caption{Hyperparameters for AlphaZero and VISA-VIS.}
  \label{tab:hyperparams}
  \centering
  \begin{tabular}{llll}
    \toprule
    \multicolumn{4}{c}{Training Parameters and  Hyperparameters}                   \\
    \cmidrule(r){1-4}
    Parameter     & Value & Value & Value     \\
    Name       & (TTT) & (4x4 TTT) & (Connect Four)     \\
    \midrule
    Num. Games & $5\mathrm{e}5$ & $1.75\mathrm{e}6$ & $7.5\mathrm{e}6$   \\
    Batch Size     & $64$ & $128$ & $256$      \\
    Learning Rate     & $1\mathrm{e}{-}3$ & $1\mathrm{e}{-}4$ & $1\mathrm{e}{-}4$      \\
    Network Depth     & $2$ & $4$ & $4$      \\
    Network Width     & $128$ & $128$ & $128$      \\
    $\lambda$     & $1\mathrm{e}{-}4$ & $1\mathrm{e}{-}4$ & $1\mathrm{e}{-}4$      \\
    \bottomrule
  \end{tabular}
\end{table}

\paragraph{Search parameters}
Recall from Section \ref{sec:alphazero} that there are a number of parameters that influence the search process of AlphaZero (and therefore VISA-VIS). During training, AlphaZero used $25$ simulations of MCTS search per time-step for Tic-Tac-Toe and 4x4 Tic-Tac-Toe, and $50$ simulations per time-step for Connect Four. For each simulation, the within-tree exploration parameter $c=2.0$ was used to guide exploration in the upper confidence bound specified by Equation \eqref{eqn:az_puct} in all games. The temperature parameter $\tau$, which guides exploration in the action selection step specified by Equation \eqref{eqn:mcts_policy}, was set to an initial value of $\tau=1.0$ for all games. During each training rollout, $\tau$ was dropped after a fixed number of steps---$5$, $9$, and $21$ for Tic-Tac-Toe, 4x4 Tic-Tac-Toe, and Connect Four, respectively---as in Silver et al. \shortcite{silver2017alphago}.
\begin{table}
  \caption{Search parameters for AlphaZero and VISA-VIS.}
  \label{tab:hyperparams}
  \centering
  \begin{tabular}{llll}
    \toprule
    \multicolumn{4}{c}{Search Parameters}                   \\
    \cmidrule(r){1-4}
    Parameter     & Value & Value & Value     \\
    Name       & (TTT) & (4x4 TTT) & (Connect Four)     \\
    \midrule
    MCTS Simulations     & $25$ & $25$ & $50$      \\
    $c$     & $2.0$ & $2.0$ & $2.0$      \\
    Initial $\tau$     & $1.0$ & $1.0$ & $1.0$      \\
    $\tau$ drop (steps)     & $5$ & $9$ & $21$      \\
    \bottomrule
  \end{tabular}
\end{table}

\paragraph{Implementation}
Our implementation is based on the open-source AlphaZero provided by DeepMind's OpenSpiel project \cite{LanctotEtAl2019OpenSpiel}. VISA-VIS extends the implementation available in OpenSpiel. The most significant implementation detail is the choice of each game's board representation. For each game, the board state was represented as a stack of $3$ $HxW$ frames---where $H$ and $W$ are the height and width of the game board, respectively---representing the pieces of each player (noughts and crosses for both Tic-Tac-Toe variants, red and yellow pieces for Connect Four) and an extra plane to encode turns (i.e. which player's turn it is to move). Each game also encoded the game rules into the action space, which are used to specify legal moves at each turn---illegal moves are masked out and assigned a probability of zero by the value network.

\paragraph{Resources}
All experiments were run on a cluster with 2x18 core Intel Xeon Skylake 6154 CPUs an Nvidia V100 GPU with 16GB of memory.

\end{document}